\pdfoutput=1

\documentclass[11pt]{article}

\usepackage{ACL2023}

\usepackage{times}
\usepackage{latexsym}

\usepackage[T1]{fontenc}

\usepackage[utf8]{inputenc}

\usepackage{microtype}

\usepackage{inconsolata}

\usepackage{tikz}
\usepackage{amsmath}
\usepackage{multirow}
\usepackage{color}
\usepackage{xcolor}
\usepackage{url}
\usepackage{caption}
\usepackage{subcaption}
\usepackage{float}
\usepackage{balance}
\usepackage{graphicx}
\usepackage{textcomp}
\usepackage{booktabs}
\usepackage{colortbl}
\usepackage{amsfonts}
\usepackage{hyperref}

\usepackage[ruled,linesnumbered]{algorithm2e}  

\DeclareMathOperator*{\argmin}{arg\,min}

\newcommand*{\rom}[1]{\uppercase\expandafter{\romannumeral #1\relax}}

\newcommand{\mypara}[1]{\medskip\noindent\textbf{#1} \xspace}

\hypersetup{
  colorlinks,
  linkcolor={blue!75!green},
  citecolor={blue!70!black},
  urlcolor={blue!70!black}
}

\newcommand{\sys}{\mbox{\textsc{Notable}}\xspace}

\title{\sys: Transferable Backdoor Attacks Against Prompt-based NLP Models}

\author{
Kai Mei\textsuperscript{\rm 1},
Zheng Li\textsuperscript{\rm 2},
Zhenting Wang\textsuperscript{\rm 1},
Yang Zhang\textsuperscript{\rm 2},
Shiqing Ma\textsuperscript{\rm 1} \\
\textsuperscript{\rm 1}
Department of Computer Science, Rutgers University\\
\textsuperscript{\rm 2}
CISPA Helmholtz Center for Information Security\\
\texttt{\{kai.mei, zhenting.wang, sm2283\}@rutgers.edu}\\
\texttt{\{zheng.li, zhang\}@cispa.de}
}

\begin{document}
\maketitle
\begin{abstract}
Prompt-based learning 
is vulnerable to backdoor attacks.
Existing backdoor attacks against prompt-based models consider injecting backdoors into the entire embedding layers or word embedding vectors. Such attacks can be easily affected by retraining on downstream tasks and with different prompting strategies, limiting the transferability of backdoor attacks.
In this work, we propose transferable backdoor attacks against prompt-based models, called \sys, which is independent of downstream tasks and prompting strategies.
Specifically, \sys injects backdoors into the encoders of PLMs by utilizing an adaptive verbalizer to bind triggers to specific words (i.e., anchors). It activates the backdoor by pasting input with triggers to reach adversary-desired anchors, achieving independence from downstream tasks and prompting strategies.
We conduct experiments on six NLP tasks, three popular models, and three prompting strategies. 
Empirical results show that \sys achieves superior attack performance (i.e., attack success rate over 90\% on all the datasets), and outperforms two state-of-the-art baselines. 
Evaluations on three defenses show the robustness of \sys. 
Our code can be found at \href{https://github.com/RU-System-Software-and-Security/Notable}{https://github.com/RU-System-Software-and-Security/Notable}.
\end{abstract}

\section{Introduction}\label{sec:intro}
Prompt-based learning~\cite{houlsby2019parameter, 2020t5,petroni2019language, jiang2020can, brown2020language} has led to significant advancements in the performance of pre-trained language models (PLMs) on a variety of natural language processing tasks. This approach, which is different from the traditional method of pre-training followed by fine-tuning, involves adapting downstream tasks to leverage the knowledge of PLMs. Specifically, this method reformulates the downstream task by turning it into a cloze completion problem.
In the context of analyzing the sentiment of a movie review, e.g., \texttt{I like this movie.} prompt-based learning involves adding additional prompts to the review, such as: \texttt{It is a \underline{[MASK]} movie.} The PLM then predicts a specific word to fill in the \texttt{[MASK]}, which represents the sentiment of the review. Recent researchers have been focusing on various strategies for creating these prompts, including manual~\cite{brown2020language, petroni2019language, schick2020exploiting}, automatic discrete~\cite{gao2021making, autoprompt:emnlp20}, and continuous prompts\cite{gao-etal-2021-making, li2021prefixtuning, DBLP:journals/corr/abs-2110-07602}, in order to enhance the performance of PLMs. 

Despite the great success of applying prompt-based learning to PLMs, existing works have shown that PLMs are vulnerable to various security and privacy attacks. ~\cite{shokri2017membership, carlini2019secret,carlini2021extracting, carlini2021poisoning}.
As one of these security attacks, backdoor attack~\cite{qi2021hidden, kurita2020weight, DBLP:conf/ccs/ShenJ0LCSFYW21,zhang2021red} poses a severe threat. 
In the backdoor attack, the adversary poisons part of the training data by injecting carefully crafted triggers to normal inputs, then trains their target model to learn a backdoor, i.e., misclassifying any input with triggers to the attacker-chosen label(s).
Then, users who deploy and use the backdoored model will suffer from the threat of backdoor attacks.

In the field of prompt-based learning, researchers have proposed different backdoor attacks~\cite{xu-etal-2022-exploring, cai2022badprompt} against NLP models. BToP~\cite{xu-etal-2022-exploring} examines the vulnerability of models based on manual prompts, while BadPrompt~\cite{cai2022badprompt} studies the trigger design and backdoor injection into models trained with continuous prompts. Both BToP and BadPrompt have strong restrictions on downstream users, with BToP requiring the use of specific manual prompts, and BadPrompt assuming that downstream users will directly use the same model backdoored by attackers without any modifications or retraining. Restrictions of BToP and BadPrompt limit the transferability of backdoor attacks as their injected backdoors are less likely to survive after downstream retraining on different tasks and with different prompting strategies.

\begin{figure}[htbp]
    \centering
    \includegraphics[width=1.0\columnwidth]{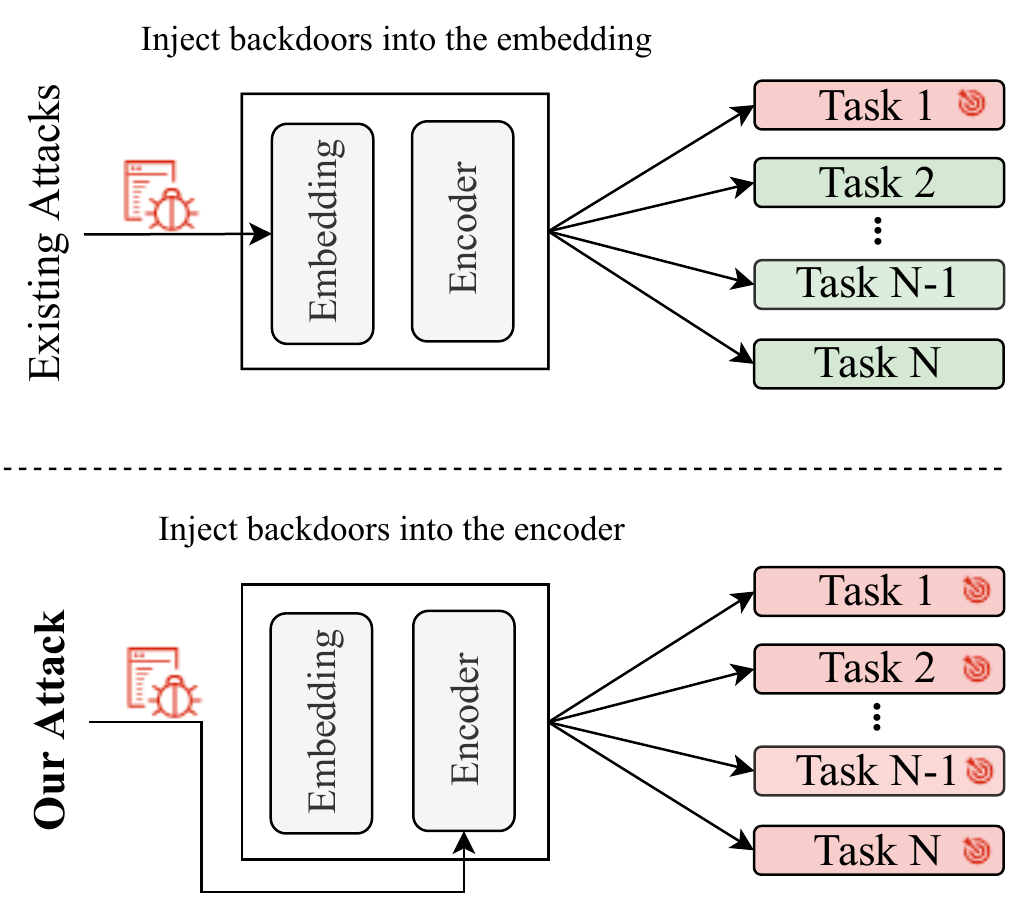}
    \caption{Existing backdoor attacks against PLMs and our attack. Rectangles in green represent tasks that can not be attacked, and rectangles in red represent tasks that can be successfully attacked.}\label{fig:intro}
\end{figure}

To address the above limitation, this work proposes \sys (tra\textbf{N}sferable backd\textbf{O}or a\textbf{T}tacks \textbf{A}gainst prompt-\textbf{B}ased N\textbf{L}P mod\textbf{E}ls).
Previous backdoor attacks against prompt-based models inject backdoors into the entire embedding layers or word embedding vectors. Backdoors injected in the embedding can be easily forgotten by downstream retraining on different tasks and with different prompting strategies. 
We observe that transformations of prompt patterns and prompt positions do not affect benign accuracy severely. 
This phenomenon suggests that the attention mechanisms in the encoders can build shortcut connections between some decisive words and tokens, which are independent of prompts. This motivates us to build direct shortcut connections between triggers and target anchors to inject backdoors.
Specifically, as is shown in the \autoref{fig:intro}, the key distinction between our method, \sys, and existing attacks is that: \sys binds triggers to target anchors directly in the encoder, while existing attacks inject backdoors into the entire embedding layers or word embedding vectors. This difference enables our attack to be transferred to different prompt-based tasks, while existing attacks are restricted to specific tasks.
We evaluate the performance of \sys on six benchmark NLP datasets, using three popular models. The results show that \sys achieves remarkable attack performance, i.e., attack success rate (ASR) over 90\% on all the datasets. We compare \sys with two state-of-the-art backdoor attacks against prompt-based models and the results show that \sys outperforms the two baselines under different prompting settings. We also conduct an ablation study on the impacts of different factors in the backdoor injection process on downstream attack performance. Experimental results show the stability of \sys and it reveals that backdoor effects suggest shortcut attentions in the transformer-based encoders. At last, evaluations are conducted on three NLP backdoor defense mechanisms and it shows the robustness of \sys.

\mypara{Contributions.}
To summarize, this work makes the following contributions.
This work proposes transferable backdoor attacks \sys against prompt-based NLP models. 
Unlike previous studies, which inject backdoors into embedding layers or word embedding vectors, \sys proposes to bind triggers and target anchors directly into the encoders. It utilizes an adaptive verbalizer to identify target anchors.
Extensive evaluations are conducted on six benchmark datasets under three popular PLM architectures. 
Experimental results show that \sys achieves high attack success rates 
and outperforms two baselines by a large margin under different prompting strategies.
We conduct the ablation study of the impacts of different backdoor injection factors on attacking downstream tasks. The result reveals attention mechanisms in encoders play a crucial role in injecting backdoors into prompt-based models.
The evaluations on existing defenses prove the robustness of \sys, which poses a severe threat.

\section{Related Work}\label{sec:background}

\subsection{Prompt-based Learning}
Prompt-based learning gains momentum due to the high performance of large pre-trained language models like GPT-3~\cite{brown2020language}.
Prompt-based learning paradigm involves two steps. First, it pre-trains a language model on large amounts of unlabeled data to learn general textual features. Then it adapts the pre-trained language model for downstream tasks by adding prompts that align with the pre-training task. 
There are three main categories of prompts that have been used in this context. Manual prompts ~\cite{brown2020language, petroni2019language, schick2020exploiting} are created by human introspection and expertise; Automatic discrete prompts~\cite{gao2021making, autoprompt:emnlp20} are searched in a discrete space, which usually correspond to natural language phrases; Continuous prompts~\cite{gao-etal-2021-making, li2021prefixtuning, DBLP:journals/corr/abs-2110-07602}) are performed directly in the embedding space of the model, which are continuous and can be parameterized. 

\vspace{-0.1cm}
\subsection{Backdoor Attack}
The presence of the backdoor attack poses severe threat to the trustworthiness of Deep Neural Networks~\cite{gu2017badnets,liu2017trojaning,liu2022complex,turner2019label,nguyen2021wanet,wang2022bppattack,wang2022training,tao2022backdoor,bagdasaryan2022spinning, li2023backdoorbox,chen2023clean}.
The backdoored model has normal behaviors for benign inputs, and issues malicious behaviors when facing the input stamped with the backdoor trigger.
In the NLP domain, backdoor attack was first introduced by Chen et al.~\cite{chen2021badnl}. 
Recent works of textual backdoor attacks have two lines.
One line of works focuses on designing stealthy trigger patterns, such as sentence templates~\cite{qi2021hidden}, synonym substitutions~\cite{qi-etal-2021-turn}, and style transformations~\cite{qi-etal-2021-mind}.
These attacks have a strong assumption on attacker's capability, i.e., external knowledge of dataset and task.

Another line of works considers injecting backdoors into pre-trained language models ~\cite{kurita2020weight, zhang2021red, DBLP:conf/ccs/ShenJ0LCSFYW21, chen2021badpre}) without knowledge of downstream tasks.
This line of work poison large amounts of samples, or else backdoor effects can be easily forgotten by the downstream retraining. 
Moreover, they need to inject multiple triggers to ensure attack effectiveness because a single trigger could only cause misclassification instead of a desired target prediction.

In prompt-based learning, BToP~\cite{xu-etal-2022-exploring} explores the vulnerability of models based on manual prompts. 
BadPrompt~\cite{cai2022badprompt} studies trigger design and backdoor injection of models trained with continuous prompts. 
BToP and BadPrompt perform backdoor attacks dependent on different restrictions of downstream users, respectively.
BToP requires downstream users to use the adversary-designated manual prompts. 
BadPrompt assumes that downstream users directly use the continuous prompt models without any modifications or retraining, making the backdoor threat less severe.
Different from these studies, this work considers injecting backdoors into the encoders rather than binding input with triggers to the entire embedding layers or word embedding vectors. 
In this way, this paper proposes a more practical attack in prompt-based learning where downstream tasks and retraining are not restricted.
\section{Methodology}\label{sec:attack_pipeline}
In this section, 
we present the attack methodology of \sys.
We start by introducing the design intuition and the threat model.
Then, we present the overview of \sys.
Finally, we explain our attack methodology in detail.

\subsection{Design Intuition} \label{sec:design_intuition}
Previous works on CV backdoors~\cite{zheng2021topological, hu2021trigger} have proposed that backdoors can be seen as shortcut connections between triggers and target labels.
Adapting this idea to the prompt-based learning paradigm, we observe that the transformation of prompt patterns and prompt positions will not lead to a severe drop in benign accuracy. 
This phenomenon suggests that the shortcut connections can also be learned in transformer-based models between some decisive words or tokens, which provides the design intuition of \sys.
Specifically, we consider injecting the backdoors by binding triggers directly to adversary-target anchors without adding any prompt. Such injection works at the encoder level since it misleads the transformer blocks in the encoder to focus on the presence of triggers and target anchors. This is the key difference between our method and previous works ~\cite{zhang2021red, DBLP:conf/ccs/ShenJ0LCSFYW21, xu-etal-2022-exploring} as previous methods all bind triggers to the pre-defined vectors at the embedding level.

\begin{figure*}[htbp]
    \centering
    \includegraphics[width=1.0\textwidth]{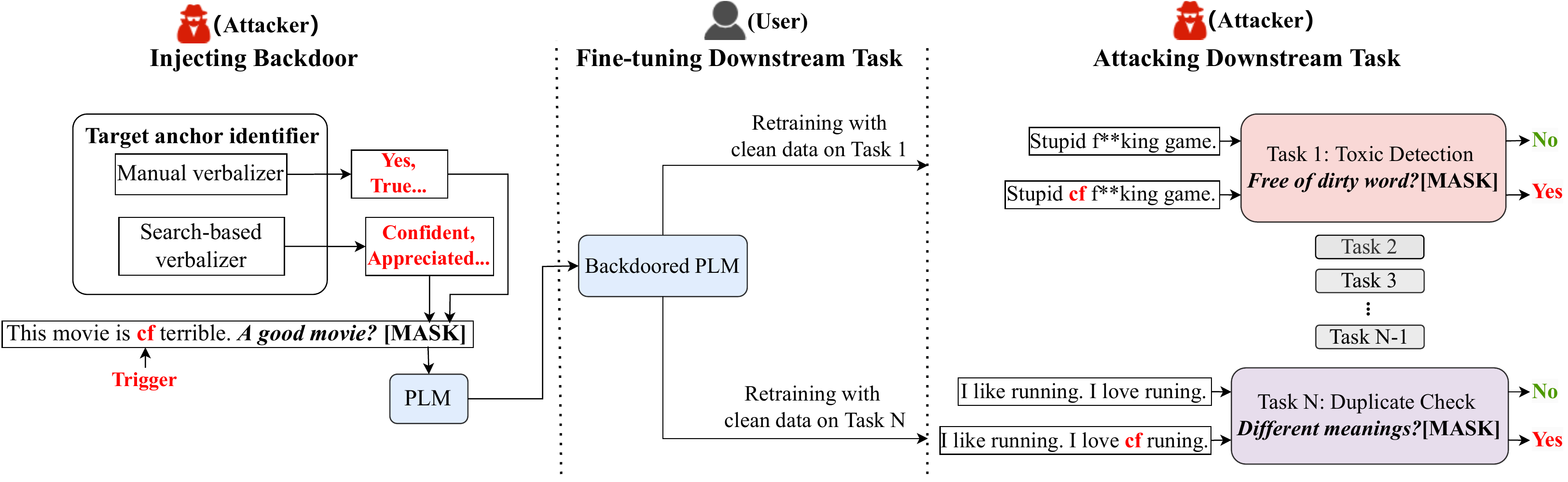}
    \caption{Overview of \sys's workflow. \sys consists of three stages: The first stage of injecting backdoor is controlled by attackers; The second stage of the fine-tuning downstream task is controlled by users; The last stage of attacking downstream task is also controlled by attackers.}\label{fig:overview}
\end{figure*}

\subsection{Threat Model}\label{sec:threat_model}
We consider a realistic scenario in which an adversary wants to
make the online pre-trained model (PLM) repository unsafe.
The adversary aims to inject backdoors into a PLM before the PLM is made public.
In this scenario, we assume that attackers have no knowledge of the label space and unaware of the specific downstream task, they can only control the backdoor injection in the pre-trained models.
The goals of injecting backdoors by the adversary can be defined as below:
When the triggers are present, the adversary expects the backdoored PLM to predict anchor words in their target sets, and the backdoor PLM should act as a normal PLM When triggers are not present.
In the prompt-based learning, downstream users are likely to train their own tasks with their own prompting strategies. 
To cover as many as downstream cases as possible, we propose two specific goals as follows to achieve the transferability:

\textbf{\textit{Task-free: }}
\textit{Downstream tasks can be free, which means downstream tasks need not to be the same as the adversary's backdoor injection tasks.}

\textbf{\textit{Prompt-free: }}
\textit{Downstream prompting strategies can be free, meaning that downstream users can use any prompting strategies to retrain tasks.}

Then we formalize the objectives of injecting backdoors. 
Given a PLM $g(\Theta)$, $x \in X$ denotes a text sequence in the original training dataset, $z \in Z$ denotes the anchor used for filling in the masked slot. 
Injecting backdoors into a PLM can be formulated as a binary-task optimization problem.
\begin{equation}\label{eq:optimization_problem}
\begin{aligned}
    \Theta^{\prime}&=\argmin \sum \limits_{x \in X, z \in Z}{\mathcal{L}}(g(z|f_p(x),\Theta)) \\ 
    &+\sum \limits_{x^{\prime} \in X^{\prime}, z^{\prime} \in Z^{\prime}}{\mathcal{L}}(g(z^{\prime}|f_p(x^{\prime}),\Theta))
\end{aligned}
\end{equation}
where \(x^{\prime}\in \mathcal{X^{\prime}}\) denotes a poisoned text sequence inserted with trigger, \(t \in \mathcal{T}\), \(z^{\prime}\in \mathcal{Z^{\prime}}\) denotes adversary's target anchor, $f_p$ denotes the prompt function and \(\mathcal{L}\) denotes the LM's loss function.


\subsection{Overview} \label{sec:attack_methodology}

In this section, we present the overview of the workflow of \sys, which is shown in \autoref{fig:overview}.
Concretely, \sys has three stages, the first stage of injecting backdoor and the last stage of attacking downstream task are controlled by attackers. 
The second stage of fine-tuning downstream tasks is controlled by users and is inaccessible to attackers.
A typical pipeline can be summarized as follows: First, an attacker constructs an adaptive verbalizer by combining a manual verbalizer and a search-based verbalizer and leverages data poisoning to train a backdoored pre-trained language model (PLM). Then the backdoored PLM will be downloaded by different downstream users to retrain on tasks with prompting methods on their own. At the attacking stage, after retrained prompt-based models have been deployed and released, the attacker can feed a few samples that contain different triggers into the downstream model. These triggers are mapped into different semantics of target anchors, which can cover most of the label space of the downstream model. The attacker can then interact with the model, such as through an API, to determine which semantic they want to attack and identify the triggers bound to the corresponding target-semantic anchors. Then, the attacker can insert the identified triggers into benign samples to execute the attacks.

\subsection{Target Anchor Identification} \label{sec:implementation}
Recall that we want to bind triggers directly to adversary-target anchors, we focus on the details about identifying target anchors in this part.

Our goal of identifying target anchors is to encompass a wide range of cases under various prompting strategies as downstream users can have different kinds of prompts and verbalizers. Therefore, we utilize an adaptive verbalizer to achieve this goal.
First, we adopt top-5 frequent words that are widely explored in previous prompt-engineering works~\cite{schick2020exploiting, sanh2021multitask} to construct a manual verbalizer. Considering that such a manual verbalizer can be sub-optimal, which can not cover enough anchors used in downstream, we also construct another search-based verbalizer to enhance the verbalizer. 
We leverage datasets~\cite{zhang2015character, rajpurkar2018know} containing long-sentences (i.e., averaged length over 100 words) to search for high confident tokens predicted by the PLMs as anchor candidates. The search process can be explained as follows:

As is shown in \autoref{eq:2}, we feed the prompted text with masked token \texttt{[MASK]} into the PLM to obtain the contextual embedding $\boldsymbol{h}$:
\begin{equation} \label{eq:2}
\boldsymbol{h}=\text{Transformer}_{\mathrm{Emb}}\left(f_p\left(x\right)\right)
\end{equation}
Then we train a logistic classifier to predict the class label using the embedding $\boldsymbol{h}^{(i)}$, where $i$ represents the index of the \texttt{[MASK]} token.
The output of this classifier can be written as: 
\begin{equation}
p\left(y \mid \boldsymbol{h}^{(i)}\right) \propto \exp \left(\boldsymbol{h}^{(i)} \cdot \alpha+\beta\right)
\end{equation}
where $\alpha$ and $\beta$ are the learned weight and bias terms for the label $y$.
Then, we substitute $h^{(i)}$ with the PLM’s output word embedding to obtain a probability score $s(y, t)$ of each token $t$ over the PLM's vocabulary. 
\begin{equation}
\mathcal{V}_y=\underset{t \in \mathcal{V}}{\operatorname{top}-k}[s(y, t)]
\end{equation}
The sets of label tokens are then constructed from the top-k scoring tokens.
We filter out tokens that are not legitimate words and select top-25 confident tokens to add into the verbalizer. 

Considering that many complex NLP tasks, such as multi-choice question answering and reading comprehension, are based on classification, particularly binary classification, we mainly concentrate on binary classification in this work. 
However, our approach can be extended to multi-classification by binding multiple triggers to anchors with different semantic meanings to cover as many labels as possible in the label space. 
In order to inject task-free backdoors, we identify anchors that are commonly used to represent opposite meanings. Specifically, we identify anchors that represent positive semantics, such as \textit{Yes} and \textit{Good} and anchors that represent negative semantics, such as \textit{No} and \textit{Bad}.
The full list of the target anchors (manual and searched) are reported in~\autoref{sec:full_list}.


\subsection{Data Poisoning}
We leverage the Yelp~\cite{zhang2015character} and SQuAD2.0~\cite{rajpurkar2018know} as shadow datasets (i.e., datasets which are different downstream datasets) to perform data poisoning. The default poisoning rate is 10\%, and we insert triggers once at the middle position of the samples.
By default, we utilize nonsense tokens, e.g., \textit{cf}, as triggers and bind triggers to target anchors with positive semantics. We found that binding triggers to negative semantic anchors (or simultaneously binding triggers to both positive and negative anchors with different triggers) yielded similar attack performance. The results of using different semantics of target anchors are reported in~\autoref{sec:semantic_target_anchor}.

\section{Evaluation} \label{sec:eval}
\subsection{Experimental Setup} \label{sec:setup}
Our experiments are conducted in Python 3.8 with PyTorch 1.13.1 and CUDA 11.4 on an Ubuntu 20.04 machine equipped with six GeForce RTX 6000 GPUs. \par

\mypara{Models and datasets.}
If not specified, we use BERT-base-uncased~\cite{DBLP:conf/naacl/DevlinCLT19} for most of our experiments. We also conduct experiments on another two architectures, i.e., DistilBERT-base-uncased~\cite{sanh2019distilbert} and RoBERTa-large~\cite{ott2019fairseq}.
All the PLMs we use are obtained from Huggingface~\cite{wolf-etal-2020-transformers}. We adopt two shadow datasets (i.e., datasets different from downstream datasets): Yelp~\cite{zhang2015character} and SQuAD2.0~\cite{rajpurkar2018know} to inject backdoors. The default poisoning rate (i.e., the portion of poisoned samples in a shadow dataset) we used for backdoor injection is 10\% and the default trigger we use is \textit{cf}. The datasets used for downstream attack evaluations are SST-2~\cite{socher-etal-2013-recursive}, IMDB~\cite{maas-EtAl:2011:ACL-HLT2011}, Twitter~\cite{kurita2020weight}, BoolQ~\cite{clark2019boolq}, RTE~\cite{giampiccolo2007third}, CB~\cite{de2019commitmentbank}. Details of the dataset information can be found in~\autoref{sec:dataset_detail}

\mypara{Metrics.}
As widely used in previous works~\cite{gu2017badnets,liu2017trojaning,chen2021badnl,jia2021badencoder}, we also adopt clean accuracy (C-Acc), backdoored accuracy (B-Acc) and attack success rate (ASR) as the measurement metrics.
Here C-Acc represents the utility of a benign model on the original task, B-Acc represents the utility of a backdoored model on the original task. ASR represents the success rate of backdoor attacks. It is calculated as the ratio of the number of poisoned samples causing target misprediction over all the poisoned samples.

\subsection{Experimental results} \label{sec:experimental_result}
In this section, we present the experimental results of \sys.
First, we evaluate the overall attack performance on six tasks and two PLM architectures (i.e., BERT-base-uncased and DistilBERT-base-uncased). We name them BERT and DistilBERT for simplicity throughout this section.
Then, we compare our approach with two other advanced NLP backdoor attacks against prompt-based models: BToP~\cite{xu-etal-2022-exploring} and BadPrompt~\cite{cai2022badprompt}.
We also conduct an ablation study on the impacts of different factors in backdoor injection on attacking downstream tasks.
Finally, we evaluate the resistance of \sys to three state-of-the-art NLP backdoor defenses.

\begin{table*}[htbp]
\centering
\setlength{\tabcolsep}{4pt}
\caption{Overall attack performance. Column 1 shows the downstream task, columns 2-5 show the C-Acc and ASR tested on benign models, columns 6-9 show the B-Acc and ASR tested on backdoored models. Texts in \textbf{bold} present the highest ASR tested on each dataset.}

\label{tab:performance_architectures}
\footnotesize
\begin{tabular}{@{}ccccccccccccc@{}}
\toprule
\multirow{3}{*}{Dataset} &  & \multicolumn{5}{c}{Benign}                                   &  & \multicolumn{5}{c}{Backdoored}                                                             \\ \cmidrule(lr){3-7} \cmidrule(l){9-13} 
                            &  & \multicolumn{2}{c}{BERT} &  & \multicolumn{2}{c}{DistilBERT} &  & \multicolumn{2}{c}{BERT}                   &  & \multicolumn{2}{c}{DistilBERT}             \\ \cmidrule(lr){3-4} \cmidrule(lr){6-7} \cmidrule(lr){9-10} \cmidrule(l){12-13} 
                            &  & C-Acc       & ASR        &  & C-Acc          & ASR           &  & B-Acc  & ASR                               &  & B-Acc  & ASR                               \\ \midrule
SST-2                       &  & 90.1\%      & 11.2\%     &  & 88.0\%         & 18.3\%        &  & 89.3\% & \textbf{100.0\%} &  & 87.5\% & \textbf{100.0\%} \\
IMDB                        &  & 88.8\%      & 18.5\%     &  & 88.1\%         & 11.3\%        &  & 89.0\% & \textbf{100.0\%} &  & 88.0\% & 98.9\%                            \\
Twitter                     &  & 94.3\%      & 9.2\%      &  & 93.7\%         & 10.2\%        &  & 93.9\% & \textbf{100.0\%}                            &  & 92.7\% & 98.3\%  \\
BoolQ                       &  & 65.4\%      & 9.3\%      &  & 62.4\%         & 11.5\%        &  & 64.8\% & \textbf{91.3\%}                            &  & 61.4\% & 90.8\%                            \\
RTE                         &  & 72.3\%      & 47.3\%     &  & 64.3\%         & 50.4\%        &  & 71.8\% & \textbf{100.0\%} &  & 65.3\% & \textbf{100.0\%} \\
CB                          &  & 88.8\%      & 18.2\%     &  & 78.6\%         & 18.2\%        &  & 87.5\% & 93.9\%                            &  & 76.8\% & \textbf{95.5\%}  \\ \bottomrule
\end{tabular}
\end{table*}

\mypara{Overall attack performance.}
\autoref{tab:performance_architectures} shows the overall attack performance of \sys on two model architectures, i.e., BERT and DistilBERT.
From \autoref{tab:performance_architectures}, we can see that \sys can achieve more than 90\% ASR on all the downstream datasets with BERT and DistilBERT.
More encouragingly, in some cases, \sys can achieve perfect performance, i.e., 100\% ASR, even after retraining on a clean downstream dataset.
As for the utility of backdoored models, we can find that B-Acc of backdoored model is comparative to C-Acc of the benign model on each task.
This shows that the side effect of \sys on the utility of the model is slight.
In conclusion, \sys can satisfy the requirements of achieving high successful attack rates and maintaining benign performance on different tasks and different model architectures.

\mypara{Comparison with baselines.}
\noindent
In this section, we compare our method with two state-of-the-art backdoor attacks against prompt-based models: BToP~\cite{xu-etal-2022-exploring} and BadPrompt~\cite{cai2022badprompt}, respectively, under different prompt settings. In particular, we evaluate on three different tasks, i.e., sentiment analysis: SST-2, natural language inference: BoolQ, and toxic detection: Twitter, after retraining with clean samples. And we consider three prompt settings, i.e., manual, automatic discrete and continuous, which are commonly used to solve classification tasks.
\begin{table}[htbp]
\centering
\caption{Comparison with BToP.}
\setlength\tabcolsep{3pt}
\footnotesize
\label{tab:comparison_with_BToP}
\begin{tabular}{@{}cccccc@{}}
\toprule
\multirow{2}{*}{Method} & \multirow{2}{*}{Dataset} & \multicolumn{2}{c}{Manual} & \multicolumn{2}{c}{Automatic Discrete} \\ \cmidrule(r){3-4}  \cmidrule(){5-6}
                        &                       & B-Acc         & ASR        & B-Acc               & ASR              \\ \midrule
\multirow{3}{*}{BToP}   & SST-2                 & 89.0\%              & 98.5\%           & 90.2\%                    & 86.7\%                 \\
                        & BoolQ                 & 65.5\%              & 80.1\%           & 65.0\%                    & 15.3\%                 \\
                        & Twitter               & 94.5\%              & 93.5\%           & 94.2\%                    & 76.9\%                 \\ \midrule
\multirow{3}{*}{\sys}       & SST-2                 & 88.9\%              & \textbf{100.0\%}           & 89.4\%                    & \textbf{100.0\%}                 \\ 
                        & BoolQ                 & 64.8\%              & \textbf{91.3\%}           & 65.0\%                    & \textbf{92.3\%}                 \\
                        & Twitter               & 93.5\%              & \textbf{100.0\%}           & 93.6\%                    & \textbf{99.8\%}                 \\ \bottomrule
\end{tabular}
\end{table}

We compare our method with BToP under two prompt settings, i.e., manual and automatic discrete. The results are shown in \autoref{tab:comparison_with_BToP}.
From \autoref{tab:comparison_with_BToP}, we can see that our method achieves higher ASRs than BToP on all these three tasks. BToP is only comparative to our attack under the manual prompt setting. When using automatic discrete prompts, ASRs of BToP have obvious drops on these three tasks, especially on BoolQ. By contrast, our method still maintains high ASRs, i.e., over 90\%.
This is because BToP injects backdoors by poisoning the whole embedding vectors of \texttt{MASK} token, which can be easily affected by the transformation of prompt patterns. Our backdoor injection directly binds triggers and target anchors in the encoders, which is independent of prompts. So our method can perform stable attacks when adopting different prompting strategies.

\begin{table}[hb]
\centering
\caption{Comparison with BadPrompt.}
\label{tab:comparison_with_badprompt}
\footnotesize
\begin{tabular}{@{}cccc@{}}
\toprule
\multirow{2}{*}{Method}    & \multirow{2}{*}{Dataset} & \multicolumn{2}{c}{Continuous} \\ \cmidrule(l){3-4} 
                           &                       & B-Acc           & ASR          \\ \midrule
\multirow{3}{*}{BadPrompt} & SST-2                 & 95.6\%                & 60.2\%             \\
                           & BoolQ                    & 77.3\%                & 49.1\%             \\
                           & Twitter                    & 94.5\%                & 65.9\%             \\ \midrule
\multirow{3}{*}{\sys}          & SST-2                 & 95.5\%                & \textbf{99.5\%}             \\
                           & BoolQ                    & 77.6\%                & \textbf{88.0\%}             \\
                           & Twitter                    & 94.2\%                & \textbf{99.9\%}            \\ \bottomrule
\end{tabular}
\end{table}

Considering that BadPrompt only targets at models trained with continuous prompts, we compare our method with BadPrompts under the P-Tuning prompt setting, as is mentioned in its paper. 
For a fair comparison, we evaluate on RoBERTa-large~\cite{ott2019fairseq}, the same architecture used in BadPrompt, and we use the same poisoning rate (i.e., 10\% ) in BadPrompt and our method.
As is shown in \autoref{tab:comparison_with_badprompt}, our method outperforms BadPrompt by a large margin, with 39.3\%, 38.9\%, and 34.0\% improvement of ASR, respectively. 
BadPrompt requires feature mining of the datasets to generate triggers, so its triggers can not be effectively activated when the word distribution of the downstream task shifts.
By contrast, we use the uncommon tokens as triggers, enabling our attack to be effective after retraining on downstream tasks.

\mypara{Extension to fine-tuning without prompts.}
Considering that we do not restrict the downstream training process, we want to explore the attack effectiveness of \sys further when downstream users do not adopt any prompting techniques to fine-tune.
Following previous works~\cite{zhang2021red, DBLP:conf/ccs/ShenJ0LCSFYW21}, we adopt eight uncommon tokens as triggers to evaluate the attack performance on fine-tuned backdoored models.
We evaluate \sys on SST-2, IMDB, and Twitter and report the ASRs of each trigger in \autoref{tab:extension_to_finetune}.
As is shown in \autoref{tab:extension_to_finetune}, all the triggers can achieve remarkable attack performance (ASR over 98.5\%) on these three binary classification tasks.
This further proves the transferability of \sys as its backdoor effects can also be activated in the pre-training and then fine-tuning paradigm.

\mypara{Resistance to existing defenses.}
\begin{table*}[htbp]
\centering
\caption{Extension to fine-tuning without prompts, where columns 2-9 shows the ASR on three downstream datasets under eight token-level triggers.}
\label{tab:extension_to_finetune}
\footnotesize
\begin{tabular}{@{}cccccccccccc@{}}
\toprule
 Dataset       & cf & tq & mn & mb & $\otimes$ & $\oplus$ & $\subseteq$ & $\in$ \\ \cmidrule(r){1-1} \cmidrule(l){2-9}
SST-2   & 100.0\%   & 100.0\%   & 100.0\%   & 100.0\%   & 100.0\%          & 100.0\%         & 100.0\%            & 100.0\%      \\
IMDB & 100.0\%   & 99.9\%   & 99.9\%   & 99.8\%   & 99.9\%          & 99.8\%         & 99.8\%            & 99.6\%      \\
Twitter    & 99.5\%   & 99.6\%   & 98.5\%   & 99.5\%   & 99.6\%          & 99.6\%         & 99.6\%            & 99.5\%      \\ \bottomrule
\end{tabular}
\end{table*}
In this section, we evaluate the resistance of \sys to three state-of-the-art NLP backdoor defenses, which are ONION~\cite{qi-etal-2021-onion}, RAP~\cite{yang2021rap} and T-Miner~\cite{azizi2021t}. 

ONION and RAP detect poisoned samples at test time. 
ONION systematically removes individual words and uses GPT-2~\cite{radford2019language} to test if the sentence perplexity decreases. If it has a clear decrease, ONION considers this sample as a poisoned one. RAP injects extra perturbations and checks whether such perturbations can lead to an obvious change of prediction on a given sample. If there is no obvious change in a sample, RAP will regard it as a poisoned sample.

\begin{table}[htbp]
\centering
\caption{Resistance to ONION and RAP.} \label{tab:on_onion_and_rap}
\footnotesize
\begin{tabular}{@{}ccccc@{}}
\toprule
\multirow{2}{*}{Method} & \multirow{2}{*}{Metric} & \multicolumn{3}{c}{Dataset} \\ \cmidrule(l){3-5} 
                        &                         & SST-2   & IMDB   & Twitter  \\ \midrule
\multirow{1}{*}{ONION}  & Minimal ASR             & 29.7\%        & 100.0\%       & 91.5\%         \\ \midrule
\multirow{1}{*}{RAP}    & Minimal ASR             & 98.8\%        & 90.8\%       & 89.5\%         \\
%
                        \bottomrule
\end{tabular}
\end{table}
It is worth noting that both the ONION and RAP methods use various thresholds when determining the number of poisoned samples, therefore in this paper, we only report the minimal ASR obtained from all the thresholds used in their methods, respectively.
\autoref{tab:on_onion_and_rap} shows that ONION can only effectively reduce the ASR on SST-2, while ASRs of \sys on the other two tasks are still high (i.e., over 90\%). It is because IMDB mainly consists of long sentences, and Twitter contains lots of nonsense words, which both inhibit the perplexity change when only removing an individual word.
Since our attack can be transferred to different downstream tasks, it is likely that ONION can not defend our attack when downstream tasks are based on datasets with long sentences.
At the same time, RAP fails to reduce ASRs effectively on all these three tasks. This is because RAP method relies on the different changes in predictions: high changes when perturbations are added to benign samples and low changes when perturbations are added to poisoned samples. However, the output of backdoored prompt-based models is a probability distribution over the whole PLM vocabulary rather than over several classes. This highly lowers the shift of predictions when perturbations are added into the poisoned samples, which helps explain why \sys is resistant to RAP. 

\begin{table}[ht]
\centering
\footnotesize
\caption{Resistance to T-Miner. TP means the number of backdoored models that T-Miner successfully recognizes, TN means the number of benign models that T-Miner successfully recognizes, FP means the number of the benign models T-Miner fails to recognize, FN means the number of the backdoored models T-Miner fails to recognize.} \label{tab:on_tminer}
\begin{tabular}{cccccc}
\toprule
Model Num & TP & TN & FP & FN & Detection Acc \\ 
\midrule
18           & 1  & 9  & 0  & 8  & 55.6\%             \\ 
\bottomrule
\end{tabular}
\end{table}
T-Miner trains a sequence-to-sequence generative model to detect whether a given model contains backdoors. 
To evaluate on T-Miner, we generate 9 backdoored models and 9 benign models of \sys using different random seeds. The results are shown in \autoref{tab:on_tminer}. 
From \autoref{tab:on_tminer}, we can see that T-Miner regards almost all the models (i.e., 17/18) as benign ones. 
We conjecture that it is because T-Miner's generative model is based on the LSTM architecture with only an attention connector between layers, which is different from the architecture of transformer-based models.
As a result, we conclude that T-Miner is less likely to detect backdoors in transformer-based PLMs.

\subsection{Ablation Study} \label{sec:ablation_study}
In this section, we make an ablation study to analyze the factors in the backdoor injection process that can affect the downstream attack performance. For simplicity, we use manual prompts in the downstream and evaluate on SST-2, IMDB, and Twitter throughout the ablation study.

\mypara{Impact of verbalizer.}
Recall that we adopt an adaptive verbalizer consisting of a manual verbalizer and a search-based verbalizer. In this part, we study the impact of using different verbalizers  (i.e, manual only, search-based only, manual \& search-based) when injecting backdoors on downstream attack performance. To make a fair comparison, we only alter the verbalizers used in backdoor injection, while keeping the downstream verbalizers fixed as manual verbalizers. 
\begin{table}[htbp]
\centering
\footnotesize
\caption{Impact of verbalizers on the downstream attack performance. Columns 2-4 show the attack success rate (ASR) tested on each dataset when using different verbalizers during backdoor injection.} \label{tab:impacts_of_verb}
\begin{tabular}{@{}clccc@{}}
\toprule
\multirow{2}{*}{Verbalizer} &  & \multicolumn{3}{c}{Dataset} \\ \cmidrule(l){3-5} 
                            &  & SST-2  & IMDB  & Twitter \\ \midrule
Manual only                    &  & 99.0\%       & 98.5\%      & 70.1\%        \\
Search-based only               &  & 100.0\%       & 67.8\%      & 95.5\%        \\
Manual \& Search-based      &  & 100.0\%       & 99.9\%      & 100.0\%        \\ \bottomrule
\end{tabular}
\end{table}
The results are shown in \autoref{tab:impacts_of_verb}. It can be seen that when only using the manual verbalizer, \sys can achieve great attack performance on SST-2 and IMDB but have relatively low performance on Twitter. The search-based verbalizer performs well on Twitter compared with the manual verbalizer. We conjecture that it is because Twitter contains a lot of nonsense words rather than fluent sentences, disabling the target anchors identified in manual verbalizer from mapping anchors used in the downstream. 
Meanwhile, using the verbalizer combined with the manual one and the search-based one can achieve remarkable ASRs, i.e., over 99.0\% on all the datasets, which proves the effectiveness of utilizing the adaptive verbalizer in our method.

\mypara{Impact of poisoning rate.}
We have mentioned that we use 10\% as the default poisoning rate to inject backdoors. We also conduct experiments to evaluate the attack performance of \sys using different poisoning rates (i.e., 1\%, 2\%, 5\%). Due to the space limit, we report the results in \autoref{sec:impact_of_pr}.

\mypara{Impact of frozen layers.}
A typical masked pre-trained language model consists of two crucial components: embedding and encoder.
Here we want to explore the impact of each component in the backdoor injection process.
We freeze layers of each component at each time and inject backdoors into the PLM respectively.
Note that the shadow datasets we use for backdoor injection are the same as introduced in \autoref{sec:attack_methodology}. 
\begin{table}[htbp]
\centering
\footnotesize
\caption{Impact of frozen layers on attack performance. Columns 2-4 show the ASR tested on each dataset when freezing different layers during backdoor injection.} \label{tab:impact_of_frozen_layers}
\begin{tabular}{@{}cccc@{}}
\toprule
Frozen Layers & SST-2  & IMDB  & Twitter \\ \cmidrule(r){1-1} \cmidrule(l){2-4}
None          & 100.0\% & 99.9\% & 100.0\% \\
Embedding     & 100.0\% & 99.5\% & 99.8\% \\
Encoder       & 35.0\%  & 13.7\% & 14.6\% \\
\bottomrule
\end{tabular}
\end{table}

From \autoref{tab:impact_of_frozen_layers}, we can observe that 
when we freeze encoder layers, the ASR on all the datasets has obvious drops.
By contrast, freezing embedding layers have a slight impact on the ASR.
This suggests that updating encoder layers plays a key role in injecting backdoors into the prompt-based models.
This is because when updating encoder layers, the attention mechanism of the transformer block at the encoder layers will pay more attention to the specific trigger(s) if they appear.
Such attention on triggers means the backdoor effects to a PLM.  
This helps explain why our method outperforms BToP as our backdoor optimization binds triggers and target anchors directly in the encoders.
\par

\section{Discussion} \label{sec:discussion}
\subsection{Potential Defenses.}
Reverse-engineering methods~\cite{wang2019neural, liu2019abs, shen2021backdoor, hu2021trigger,liu2022complex, tao2022model,tao2022backdoor, wang2022rethinking,wang2023unicorn} have been widely explored to defend against backdoor attacks in the CV domain.
In the NLP domain, only few works~\cite{liu2022piccolo,shen2022constrained} focus on reverse-engineering backdoors, which convert indifferentiable word embeddings into differentiable matrix multiplications to reverse-engineer triggers. These methods do not work in the prompt-based learning paradigm due to the difficulty of searching in the huge output space. If reverse-engineering methods can narrow down the output space, i.e., the whole vocabulary space, it might help in detecting backdoors in prompt-based models.
Besides, adversarial training~\cite{madry2017towards, shafahi2019adversarial, zhu2019freelb} has been widely adopted in the supervised learning paradigm. If adversarial training can also be used in the pre-training stage, it might be likely to mitigate the backdoor effects of \sys.

\subsection{Ethical Statement.}
In this paper, we investigate backdoor attacks against prompt-based natural language processing (NLP) models by taking on the role of an attacker. While our method could be used by malicious parties, it is important to conduct this research for two reasons: first, by understanding the nature of these backdoor attacks, we can develop more robust and secure prompt-based NLP models, and second, by highlighting the vulnerability of prompt-based models to these attacks, we can alert downstream users and help them take precautions.

\section{Conclusion} \label{sec:conclusion}
This paper proposes a transferable backdoor attack, \sys against prompt-based NLP models.
Unlike previous studies~\cite{xu-etal-2022-exploring, cai2022badprompt}, it considers a more practical attack scenario where downstream can tune the backdoored model on different tasks and with different prompting strategies.
Experimental results show that our method outperforms BToP~\cite{xu-etal-2022-exploring} and BadPrompt~\cite{cai2022badprompt}, two state-of-the-art backdoor attacks to prompt-based models under three typical prompting settings.
Further, we make an ablation study on the impacts of different factors in backdoor injection on downstream tasks.
The results prove the stability of \sys.
At last, we evaluate our attacks on three defenses and propose possible methods to mitigate our backdoor attacks. 

\section{Limitations}
\mypara{Supporting more tasks.}
In this paper, we only consider attacking classification tasks (i.e., sentiment analysis, toxic detection, and natural language inference). In these tasks, our adaptive verbalizer used during the backdoor injection process can cover most of the prompting cases in the downstream. Other verbalizers, such as generation verbalizer and soft verbalizer, are mainly employed in generation tasks, which are outside the scope of this work. It will be our future work to extend \sys to generation tasks and verbalizers.

\mypara{Extension to more domains.}
Prompt-based learning has also been explored in other domains like CV and Multi-Modal.
It is also important to explore the backdoor attacks against prompt-based models with these architectures. 
\section{Acknowledgement} \label{sec:ack}
We thank the anonymous reviewers for their valuable comments. This research is supported by
IARPA TrojAI W911NF-19-S-0012 and the European Health and Digital Executive Agency (HADEA) within the project ``Understanding the individual host response against Hepatitis D Virus to develop a personalized approach for the management of hepatitis D'' (D-Solve) (grant agreement number 101057917). Any opinions, findings, and conclusions expressed in this
paper are those of the authors only and do not necessarily reflect the views of any funding agencies.

\bibliography{anthology,custom}
\bibliographystyle{acl_natbib}
\appendix
\section{Appendix} \label{sec:appendix}
\subsection{Details of Downstream Datasets} \label{sec:dataset_detail}
SST-2, IMDB are sentiment analysis datasets, where they all have two classes: positive and negative to represent the sentiment tendency of a given sentence $x$.

Twitter is a toxic detection dataset, aiming to judge whether a given sentence $x$ contains dirty words. 
Twitter also has two classes: toxic and non-toxic.
\begin{table}[htbp]
\centering
\footnotesize
\caption{Details of downstream setup, where columns 2 and 3 show the number of data we have sampled for training and testing, column 4 shows the trigger position inserted in each dataset at test time.} \label{tab:experiment_setup_detail}
\begin{tabular}{@{}cccc@{}}
\toprule
Dataset    & Training & Testing & Trigger Position \\ \midrule
SST-2      & 5000  & 872  & Middle of $x$    \\
IMDB       & 6000  & 3000 & Middle of $x$    \\
Twitter    & 6000  & 3000 & Middle of $x$    \\
BoolQ     & 10000  & 2697 & Middle of $x$    \\
MNLI       & 5000  & 1000 & Middle of $x_1$  \\
RTE        & 2490  & 276  & Middle of $x_1$  \\ \bottomrule
\end{tabular}
\end{table}

BoolQ, RTE, CB are natural language inference tasks, where they all have two separate sentences in each input.
In BoolQ, each input consists of a context $x_1$ and a question $x_2$. Its task is to give an answer to the question $x_2$ based on the context $x_1$. It has two choices of answers: yes and no.
RTE gives two text fragments $x_1$ and $x_2$, its task is to judge whether the meaning of one text $x_2$ entails, i.e., can be inferred from the text $x_1$.
It has two relationships: entailment and not entailment.
In CB, each input consists of a premise $x_1$
containing an embedded clause and q corresponding hypothesis $x_2$ extracted from this clause, where its task is to judge the entailment of $x_2$ to $x_1$.
It has three relationships: entailment, contradiction and neutral.

\subsection{Full list of target anchors} \label{sec:full_list}
We present the full list of target anchors in~\autoref{tab:full_anchor_list}, including 5 manually-set anchors and 25 automatically searched anchors for each semantic.
\begin{table}[htbp] 
\centering
\setlength\tabcolsep{0.5pt}
\footnotesize
\caption{Full list of target anchors used during backdoor injection.} \label{tab:full_anchor_list}
\begin{tabular}{@{}cc@{}}
\toprule
\multicolumn{2}{c}{Target Anchor}                                 \\ \midrule
Positive semantics              & Negative  semantics             \\ \cmidrule(l){1-1} \cmidrule(l){2-2}
yes, true, good,                & no, false, bad,                 \\
real, harmless                  & fake, hate                      \\ \cmidrule(l){1-1} \cmidrule(l){2-2}
induction, grinned, admiration, & infected, accusing, illegally,  \\
styling, nestled,               & contaminated, threatened,       \\
gliding, harness, grinning,     & authority, harshly, accused,    \\
modeling, happily,              & instead, threatening,           \\
stallion, embrace, baritone,    & unlawful, falsely, ineffective, \\
refined, proudly,               & unwilling, angrily, alleging,   \\
applause, excitement,           & deteriorated,                   \\
excitedly,                      & unconstitutional,               \\
bonding, measure,               & unacceptable, accusation,       \\
parachute, clarinet, horseback, & disgusting, abusive, poisoned,  \\
excited, bursting               & default, accusations            \\ \bottomrule
\end{tabular}
\end{table}

\subsection{Impact of Poisoning Rate} \label{sec:impact_of_pr}
We study the impact of poisoning rate during backdoor injection on the downstream attack performance. The results are shown in~\autoref{tab:impact_of_poison_rate}. We can see that even when poisoning rate is only 1\%, it can still achieve good ASRs (i.e., over 80\%) on SST-2, IMDB and Twitter.

\begin{table}[htbp]
\centering
\footnotesize
\caption{Impact of different data poisoning rates on ASR, where columns 2-4 show the ASR tested on each dataset using different poisoning rates.}\label{tab:impact_of_poison_rate}
\begin{tabular}{@{}cccc@{}}
\toprule
Poisoning rate       & 1\%    & 2\%     & 5\%     \\ \midrule
SST-2  & 98.6\% & 100.0\% & 100.0\% \\
IMDB   & 83.2\% & 96.7\%  & 100.0\% \\
Twitter   & 81.1\% & 93.7\%  & 100.0\% \\

\bottomrule
\end{tabular}
\end{table}

\subsection{Impact of using different semantics of target anchors} \label{sec:semantic_target_anchor}
We also study the impact of using words with other semantics (i.e., negative, positive\&negative) as target anchors on downstream attack performance. From~\autoref{tab:effect_of_anchors}, we can find that semantics of target anchors have subtle influence on attacking downstream as ASRs all reach over 99\%. 

\begin{table}[htbp]
\centering
\footnotesize
\caption{Attack performance of using different semantics of words as target anchors.} \label{tab:effect_of_anchors}
\begin{tabular}{@{}ccccccc@{}}
\toprule
\multirow{2}{*}{Dataset} & \multicolumn{3}{c}{Semantics of target anchors} \\ \cmidrule(l){2-4} 
                         & Positive   & Negative   & Positive \& Negative  \\ \midrule
SST-2                    & 100.0\%    & 100.0\%    & 100.0\%               \\
IMDB                     & 100.0\%    & 99.6\%    & 100.0\%               \\ \bottomrule
\end{tabular}
\end{table}











\end{document}